\documentclass[10pt,twocolumn,letterpaper]{article}

\usepackage{cvpr}
\usepackage{times}
\usepackage{epsfig}
\usepackage{graphicx}
\usepackage{amsmath}
\usepackage{amssymb}
\usepackage{subcaption, booktabs}
\usepackage{xcolor}
\usepackage{multirow}

\graphicspath{ {./figures/} }

\usepackage[pagebackref=true,breaklinks=true,letterpaper=true,colorlinks,bookmarks=false]{hyperref}

\cvprfinalcopy % *** Uncomment this line for the final submission

\def\cvprPaperID{****} % *** Enter the EV Paper ID here

\ifcvprfinal\fi
\begin{document}

%%%%%%%%% TITLE
\title{An Attention-Based System for Damage  \\Assessment Using Satellite Imagery}

\author{\parbox{16cm}{\centering
    {Hanxiang Hao$^{\star}$ \quad Sriram Baireddy$^{\star}$ \quad Emily R. Bartusiak$^{\star}$ \\ Latisha Konz$^{\dagger}$ \quad Kevin LaTourette$^{\dagger}$  \quad Michael Gribbons$^{\dagger}$ \quad Moses Chan$^{\dagger}$ \\ Mary L. Comer$^{\star}$ \quad Edward J. Delp$^{\star}$}\\
    {\normalsize
    $^{\star}$ Video and Image Processing Lab (VIPER), Purdue University, West Lafayette, Indiana USA\\
    $^{\dagger}$ Lockheed Martin Space,
    Sunnyvale, California USA}}
}
\maketitle
%\thispagestyle{empty}

%%%%%%%%% ABSTRACT
\begin{abstract}
When disaster strikes, accurate situational information and a fast, effective response are critical to save lives.
Widely available, high resolution satellite images enable emergency responders to estimate locations, causes, and severity of damage.
Quickly and accurately analyzing the extensive amount of satellite imagery available, though, requires an automatic approach. 
In this paper, we present Siam-U-Net-Attn model -- a multi-class deep learning model with an attention mechanism -- to assess damage levels of buildings given a pair of satellite images depicting a scene before and after a disaster.
We evaluate the proposed method on xView2, a large-scale building damage assessment dataset, and demonstrate that the proposed approach achieves accurate damage scale classification and building segmentation results simultaneously.
\end{abstract}

%%%%%%%%% BODY TEXT
\section{Introduction}

\label{sec:intro}
\begin{figure}[t]
  \begin{center}
    \includegraphics[width=0.8\linewidth]{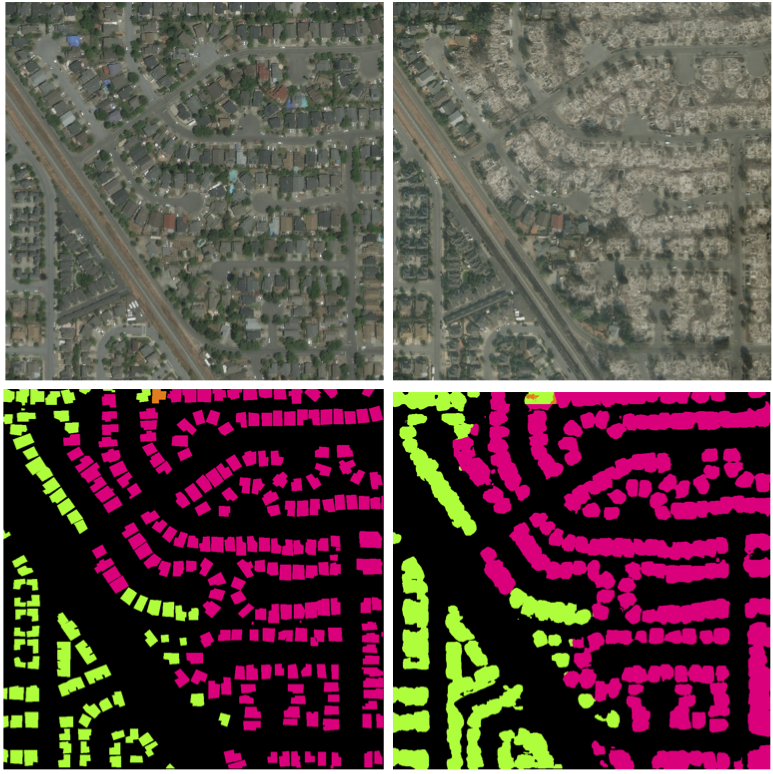}  
  \end{center}
    \caption{\textbf{Damage Scale Classification Components}. From left to right and top to bottom: pre-disaster input image, post-disaster input image, ground truth mask, and damage scale classification output mask. The green areas illustrate buildings with \textit{no damage}, and the pink areas reveal \textit{destroyed} buildings.}
  \label{fig:classification}
\end{figure}

Natural disasters wreak havoc on nations.
They kill approximately 90,000 people every year and affect 160 million people around the globe~\cite{WHO_2020}.
Furthermore, areas afflicted by weather and climate disasters sustain significant physical, social, and economic devastation.
Short-term effects of disasters evolve into long-term ramifications that linger for years to come~\cite{Boustan_2017, WHO_2020}.
Considering economic consequences alone reveals staggering figures.
For example, the 2010 Haiti earthquake inflicted approximately \$7.8 billion - \$8.5 billion in damages to infrastructure~\cite{Amadeo_2019}.
In 2019, the United States endured fourteen distinct natural disasters whose overall damages each exceeded \$1 billion dollars~\cite{NOAA_2020}.
Environmental climate analyses also indicate that the frequency and brutality of natural disasters will increase in the future due to climate change and rising greenhouse gas emissions~\cite{Aalst_2006, Boustan_2017}.
Therefore, the impact of disasters is immediate, far-reaching, and continuous growing.
% The United States has already experienced an increase in the occurrence of disasters.
% Over the last 40 years, the U.S. suffered an average of 6.5 disasters every year, each costing over \$1 billion.
% However, in the last five years, the U.S. experienced an increased average of 13.8 severe, billion-dollar disasters annually~\cite{NOAA_2020}.

With the increase in severity and regularity of disasters, preparation for disaster recovery and emergency resource planning is needed now more than ever.
Emergency responders require rapid and reliable situational details to save disaster victims while ensuring their own safety during rescue efforts.
Moreover, accurate damage estimates assist responders in determining evacuation plans and in preventing secondary disasters caused by collapses of damaged buildings. 
In the long run, damage assessment estimates also empower planning efforts for building and infrastructure repairs.

Very high resolution (VHR) satellite imagery is increasingly available due to an ever-expanding fleet of commercial satellites, such as DigitalGlobe's WorldView satellites~\cite{Dg_2020}.
VHR imagery enables detailed assessment of disaster damage at the building level. 
With the recent improvement in machine learning methods, especially deep learning approaches, rapid analysis of large amounts of VHR satellite imagery is feasible and this facilitates damage estimation and aids in disaster relief efforts. 
% More importantly, these machine learning tools enable fast emergency responses to save lives and reduce damage costs. 
In this paper, we propose a Siam-U-Net-Attn model to quickly and accurately estimate the damage of a disaster.
Our approach analyzes two satellite images of the same scene, acquired before and after a disaster. 
It then produces a mask showing buildings with labels that indicate different damage scale levels, as depicted in Figure \ref{fig:classification}.

The main contributions of this work include:
\begin{itemize}
    \item Development of a multi-class deep learning model with attention technique that accurately classifies damage levels of buildings in satellite imagery. 
    \item Production of semantic building segmentation masks using the proposed method.
    \item Demonstration that the proposed model achieves better results for building damage scale classification than other methods while simultaneously achieving accurate building segmentation results. 
\end{itemize}

\section{Related Work}
\label{sec:related_work}

The proposed method achieves building damage scale classification by analyzing buildings within satellite imagery and determining the level of damage inflicted to them.
Due to limited amounts of labeled data, most research addressing damage scale classification instead simplifies this multi-class task to a change detection operation, which assigns a binary label, \textit{damage} or \textit{no-damage} to each building.
Research approaches for solving change detection fall into several broad categories~\cite{Asokan_2019}.

Algebra-based change detection techniques perform mathematical operations on image pixels to obtain a difference image.
Such approaches, including image differencing~\cite{Ke_2018} and change vector analysis~\cite{Qi_2015}, involve a threshold selection process to determine which components changed.
Algebra-based change detection methods are relatively simple to implement, but they do not provide contextual information about the detected changes.

Transform-based change detection approaches transform event images. 
Image transforms, including the standard principal component analysis approach~\cite{Sadeghi_2016}, strive to determine pertinent information for the change detection task.
While transforming the images enables analysis of change in a different dimensionality, it also presents challenges in labeling regions of change in the event images themselves.

Classification-based change detection methods usually rely on larger amounts of labeled data. 
They easily extend to the multi-class damage scale classification task considered in this paper.
Xu \etal~\cite{Xu_2019} and Fujita \etal~\cite{Fujita_2017} describe several models for this objective, including a single-stream model and a double-stream model (\ie, Siamese network).
Their models evaluate a pair of input images of a scene before and after a disaster.
Then, they produce a single binary classification label, indicating whether the image contains \textit{damage} or \textit{no-damage}.
Similarly, Nex \etal~\cite{Nex_2019} propose a binary classification model based on DenseNet~\cite{Huang_2017} with dilated convolution~\cite{Yu_2016} to achieve a larger receptive field.
Mou \etal~\cite{Mou_2019} and Lyu \etal~\cite{Lyu_2016} introduce recurrent neural networks to jointly learn spectral-spatial-temporal features for change detection. 
Connors \etal~\cite{Connors_2017} design a semi-supervised method that uses a variational autoencoder~\cite{Kingma_2014} to infer change detection labels without ground truth for every training instance.
An unsupervised method is proposed by Liu \etal~\cite{Liu_2019} using active learning~\cite{Wang_2017} to construct training samples and using graph convolutional network~\cite{Garcia_2018} for change detection.
However, none of these approaches produce pixel-wise classification masks.

There is some research in constructing building classification masks in an unsupervised manner.
Jong \etal~\cite{Jong_2019} utilize the U-Net model~\cite{Ronneberger_2015} to detect changes in satellite images.
They first train a U-Net model for the building segmentation tasks.
During change detection inferencing, they collect two sets of features from the trained U-Net model (\ie, activations of different layers in the U-Net), given two query images.
Then, the difference of the two sets of features forms the change detection map.
Another approach is a deep convolutional coupling network proposed by Liu \etal~\cite{Liu_2018} uses both optical and radar images for unsupervised change detection.  
They use an ad-hoc weight initialization for the network that is based on the noise models of the optical and radar images to help the model learn the proper features during training.

\begin{figure*}[t]
  \begin{center}
    \includegraphics[width=0.7\linewidth]{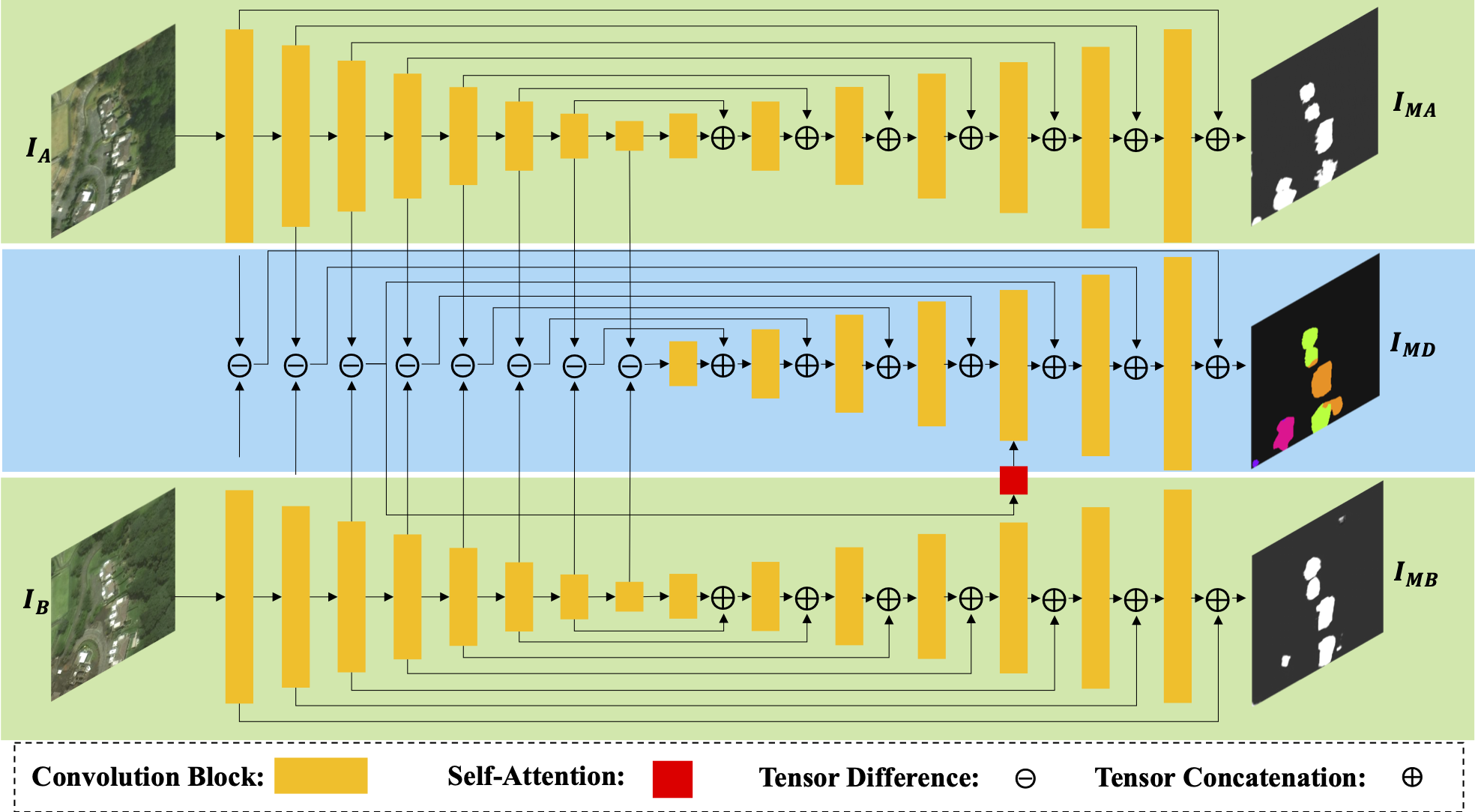}  
  \end{center}
    \caption{\textbf{Architecture of Proposed Method:  Siam-U-Net-Attn-diff model}. $I_A$ and $I_B$ are the pre-disaster and post-disaster input images. $I_{MA}$ and $I_{MB}$ are the corresponding output building segmentation masks. $I_{MD}$ is the output damage scale classification mask.}
  \label{fig:method}
\end{figure*}

Supervised classification methods constitute the final category of solutions to the change detection task.
Demir \etal~\cite{Demir_2013} propose a method that only requires the annotation of one image in a time series. 
They train a supervised classification model using a dataset constructed by an active learning approach~\cite{Wang_2017}.
Chu \etal~\cite{Chu_2016} apply deep belief networks (DBNs)~\cite{Hinton_2006} to produce a change detection map.
Two DBNs are used for extracting features from the image regions that contain changes and do not contain changes, respectively. 
They compare the feature distances obtained from the two DBNs for each image patch to construct the change detection map.
Papadomanolaki \etal~\cite{Papadomanolaki_2019} combine the U-Net model with a LSTM~\cite{Hochreiter_1997} model in order to use temporal information from multiple frames of satellite imagery.
Compared to results that use only two input frames, their model achieves better performance.
Daudt \etal~\cite{Daudt_2018} propose using an encoder-decoder-based architecture to produce the change detection map.
The decoder upsamples features extracted from the encoder to generate a mask indicating damage levels throughout the region under analysis.
They also improve on this performance in~\cite{Daudt_2019} by combining the semantic segmentation task with the change detection task to achieve multi-task learning. 
They use two U-Net models in total; one for each task.
The semantic segmentation U-Net utilizes one image (taken either before or after the change event) to produce the segmentation mask of objects of interest. 
The change detection U-Net utilizes two images (\ie, one from before the changes and one from after the changes) as well as the features extracted from the semantic segmentation model to produce the change detection mask. 
By fusing the features together, they achieve better performance in the change detection task.

Our proposed model extends this concept and combines the previously mentioned U-Net model with the Siamese model.
The U-Net model learns the semantic segmentation of buildings, while the Siamese model learns the damage scale classification.
The use of the Siamese model allows us to reduce the number of learned parameters and the size of the model during both training and inferencing in comparison to~\cite{Daudt_2019}.
By combining these models, we achieve multi-task learning of both segmentation and classification.
Additionally, we introduce a self-attention module that improves the performance by incorporating long-range information from the entire image.

\section{Our Proposed Method}
\label{sec:method}

We propose a Siam-U-Net-Attn model for damage classification and building segmentation, as shown in Figure \ref{fig:method}.
It is inspired by~\cite{Daudt_2018, Ronneberger_2015}. 
One element of this architecture is a U-Net model that analyzes a single input image and produces a segmentation mask showing building locations in the input image.
The U-Net model is a fully convolutional network that was proposed by~\cite{Ronneberger_2015} for image segmentation.
Besides its encoder-decoder structure for local information extraction, it also utilizes skip connections to retain global information.
A single U-Net model analyzes input frames $I_A$ and $I_B$, which depict the same scene pre-disaster and post-disaster, respectively. 
Since the U-Net focuses on the building segmentation objective, it is agnostic to the disaster.
In other words, we can use the same model for both pre-disaster and post-disaster images to produce binary masks $I_{MA}$ and $I_{MB}$, corresponding to their respective input frames.
The two green regions in Figure \ref{fig:method} indicate the shared U-Net model for $I_A$ and $I_B$.

The features extracted from the encoder regions of the U-Net model also assist in the damage scale classification task.
The two-stream features produced by the U-Net encoder and a new, separate decoder constitute the Siamese network, shown as the blue region in Figure \ref{fig:method}.
In the Siamese network, we compare features from the two input frames to detect the damage levels of buildings.
Simple differencing and channel-wise concatenation are two methods to compare the two-stream features.
By comparing features from the two frames, the Siamese model evaluates the differences between the features in order to assess the damage levels.
Figure \ref{fig:method} shows the architecture of the Siam-U-Net-Attn in difference mode (\ie, Siam-U-Net-Attn-diff).
The Siam-U-Net-Attn in concatenation mode (\ie, Siam-U-Net-Attn-conc) can be obtained by replacing the difference operations with channel-wise concatenation operations. 
In Section \ref{sec:experiment}, we will compare the performance of the proposed model in difference and concatenation modes.

\begin{figure}[t]
  \begin{center}
    \includegraphics[width=0.7\linewidth]{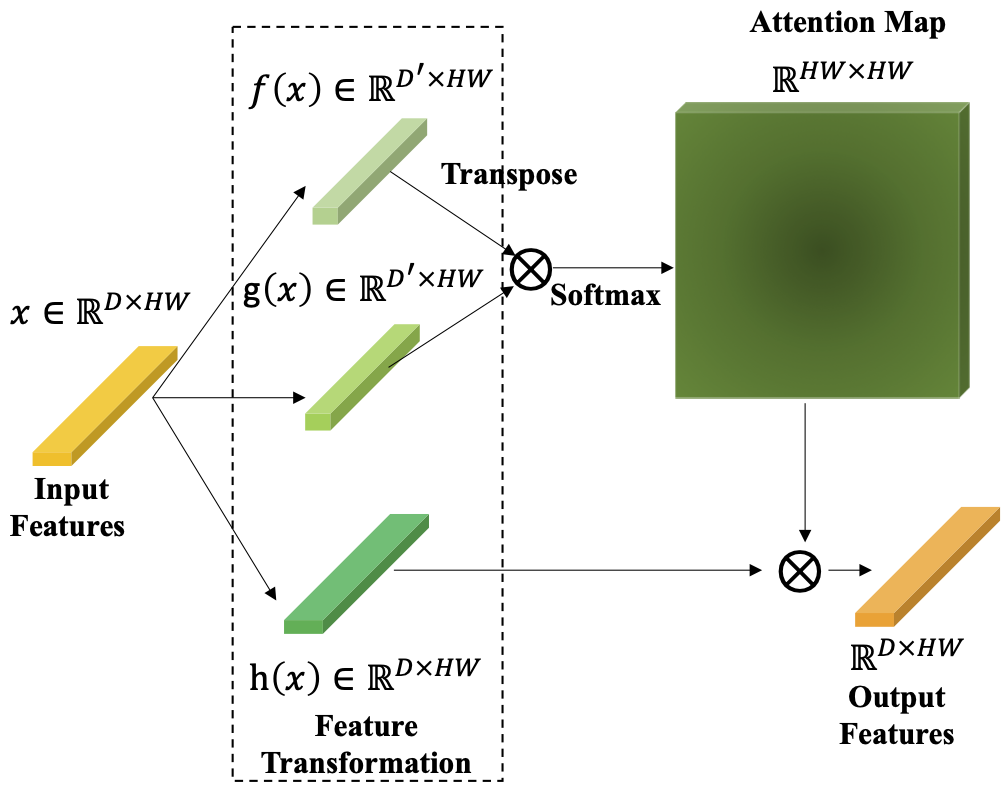}  
  \end{center}
    \caption{\textbf{Architecture of the Self-Attention Module}. This modified figure is based on~\cite{zhang_2019}.}
  \label{fig:attention_method}
\end{figure}

Analyzing a building by itself is not sufficient for accurate damage level classification.
It is also necessary for the network to consider the area surrounding buildings in its assessment.
For example, natural disasters such as floods may not damage a building's roof, but water surrounding the building may indicate interior damage.
% Likewise, volcanic flow surrounding a building is an indication of damage, even if the building itself has no visible damage.
Since convolution is a local operation that can only access local neighborhoods, we use a self-attention module~\cite{Wang_2018, zhang_2019} to capture long-range information.
Figure \ref{fig:attention_method} illustrates the mechanism of the self-attention module.
Assume the input feature map is $\mathbf{x}\in\mathbb{R}^{D \times N}$, where $N$ is the flattened size of feature map along the height and width dimensions (\ie $N=H \times W$) and $D$ is the number of channels of the input feature.
To compute the attention map, we first transform the input features into two feature spaces by:
\begin{align*}
  \mathbf{f}(\mathbf{x}) = \mathbf{W_f}\mathbf{x}, \ \ \ \mathbf{g}(\mathbf{x}) &= \mathbf{W_g}\mathbf{x}.
\end{align*}
The attention map is calculated as
\begin{align*}
  \mathbf{a}(\mathbf{x})=\text{Softmax}(\mathbf{f}(\mathbf{x})^T \mathbf{g}(\mathbf{x})).
\end{align*}
The $\text{Softmax}$ function is computed along the second dimension to normalize each row of the attention map.
We then apply the attention map to the input features as: 
\begin{align*}
  \mathbf{o}(\mathbf{x})=\mathbf{h}(\mathbf{x})\mathbf{a}(\mathbf{x})^T,
\end{align*}
where $\mathbf{h}(\mathbf{x}) = \mathbf{W_h}\mathbf{x}$.

$\mathbf{W_f}\in\mathbb{R}^{D' \times D}$, $\mathbf{W_g}\in\mathbb{R}^{D' \times D}$, and $\mathbf{W_h}\in\mathbb{R}^{D \times D}$ are trainable parameters that are implemented as the convolution operation with a kernel size of $1\times1$.
According to~\cite{zhang_2019}, we choose $D' = D/8$ to reduce memory usage.
The final output of the self-attention module is a weighted summation of the original input with the attention feature:
\begin{align*}
  \mathbf{y}(\mathbf{x})=\gamma\mathbf{o}(\mathbf{x})+\mathbf{x},
\end{align*}
where $\gamma \in \mathbb{R}$ is also a learnable parameter.
Therefore, each value of the self-attention output contains information of every input feature provided by the attention map.
As shown in Figure \ref{fig:method}, the model invokes a self-attention module after merging the features from the two input frames.
It is important to note that the attention map from the self-attention module requires a lot of memory for large-resolution features, so we place the module in a low resolution layer of size $32\times32$ to reduce the memory usage.

\section{Dataset}
\label{sec:dataset}

In this paper, we use the xView2 dataset~\cite{Gupta_2019} for both training and testing. 
This dataset is designed for the task of building damage assessment and covers a wide variety of disaster events, such as tsunamis, earthquakes, and volcanic eruptions.
It contains 2,799 pairs of pre-event/post-event multi-band images with resolution $1024\times1024$ pixels.
Additionally, it contains segmentation ground truth masks with building polygons and classification labels indicating damage levels.
There are four damage levels: \textit{no-damage}, \textit{minor-damage}, \textit{major-damage}, and \textit{destroyed}.
~\cite{Gupta_2019} describes the scoring method used to assign damage levels.

To reduce the memory usage during training and testing, we use image patches of size $256\times256$ as the inputs to our system.
We crop every satellite image into 16 non-overlapping patches, each sized $256\times256$.
The final dataset contains 44,784 pairs of image patches.
We also use data augmentation methods (\ie, horizontal/vertical flipping, random color jittering, and random cropping) during training to reduce overfitting.
Random color jittering and cropping are applied independently to pre-event and post-event images to simulate poor image normalization and registration.
We implement two different data splitting methods to separate the dataset into training, validation and testing sets.
For the first split (\textit{Split I}), we crop full-resolution images into patches and then separate the patches into training, validation, and testing sets according to a ratio of $0.6:0.2:0.2$.
For the second split (\textit{Split II}), we separate the full-resolution images into the different sets before cropping them into patches.
The reason for these two dataset splits is to explore how the method performs on scenes it has never seen before. 
In \textit{Split I}, the training and testing datasets could both contain image patches from the same full-resolution image.
Thus, patches of the same scene could be contained in both the training and testing sets.
In \textit{Split I}, we ensure that the training and testing patches come from different full-resolution images.
Therefore, the training and testing datasets contain different scenes, simulating performance in a real-world scenario when the model is presented with images it has never seen.

\section{Experimental Results}
\label{sec:experiment}

As shown in Figure \ref{fig:method}, our model consists of eight convolution blocks for the encoder and decoder components.
We design the eight convolution blocks to ensure the resolution of the middle layer (\ie, the layer with the smallest feature resolution) is $1\times1$. 
Each downsampling block consists of convolution, ReLU, batch normalization, and maxpooling layers.
Each upsampling block consists of upsampling with bilinear interpolation, convolution, batch normalization, and ReLU layers. 
The output damage scale classification mask has five channels: the four damage levels plus one \textit{background} label.
We use weighted binary cross-entropy loss and multi-label cross-entropy loss for the building segmentation loss $\mathcal{L}_s$ and damage scale classification loss $\mathcal{L}_d$, respectively, which are defined as:
\begin{align*}
  \mathcal{L}_s & = -(w_{s,1}y_s\log{p_s} + w_{s,0}(1-y_s)\log{(1-p_s)}) \\
  \mathcal{L}_d & = -\sum_{c=1}^{5}{w_{d,c}y_d(c)\log{p_{d}(c)}}
\end{align*}
$y_s$ and $p_s$ are the ground truth label and the detected building segmentation probability, respectively, while $y_d(c)$ and $p_d(c)$ are the ground truth label and the detected classification probability for damage scale $c$. 
$w_s$ and $w_d$ are weights applied to each class to address the class imbalances present in our dataset.
Since most areas in our images do not contain any buildings, we choose a larger weight for the \textit{building} class, indicated by $w_{s,1}$ in the segmentation loss $\mathcal{L}_s$.
Additionally, undamaged buildings are more common than damaged buildings in our dataset. 
Therefore, we also select larger weights for the damaged-building classes ($c = 2, 3, 4$) compared to the non-damaged buildings ($c = 1$) in the damage scale classification loss $\mathcal{L}_d$.
Table \ref{table:weights} shows the empirical weights we institute.
\begin{table}[t]
	\begin{center}
		\begin{tabular}{cccccc}
			\toprule
      \multirow{2}{*}{\textbf{Weights}} & \multicolumn{5}{c}{\textbf{Label}}  \\
              & 0 & 1  &  2 &  3 & 4  \\ \midrule
      $w_s$   & 1 & 10 & -  & -  & -  \\
      $w_d$   & 1 & 10 & 30 & 30 & 30 \\
			\bottomrule
		\end{tabular}
		\caption{\textbf{Class Balancing Weights.} Weights of the binary cross entropy loss and multi-label cross entropy loss for the imbalanced building segmentation and damage scale classification tasks.}
		\label{table:weights}	
	\end{center}
\end{table}
The final loss function is the summation of the two building segmentation losses for the two input frames $I_A$ and $I_B$ plus the damage scale classification loss.
The Adam optimizer~\cite{Kingma_2015} is used to train the proposed models.
We train our model for 100 epochs with an initial learning rate of 0.001.
The learning rate linearly decays to zero in the final epoch.

We compare the two proposed models with the three methods from~\cite{Daudt_2018}: fully convolutional early fusion (FC-EF), fully convolutional Siamese-difference (FC-Siam-diff), and fully convolutional Siamese-concatenation (FC-Siam-conc).
The FC-EF model is essentially the U-Net model we described previously.
Its input is $I_A$ and $I_B$ after concatenation along their channels.
The FC-Siam-diff and FC-Siam-conc models utilize the Siamese model without the U-Net decoder used in the proposed method.
These methods are designed for the change detection task and thus operate in a binary classification fashion.
To compare these models with our proposed method, we changed their output layers from binary classification layers to multi-label classification layers.
We also used the same training settings we selected for our method, including the optimizer and learning rate, since the authors of the compared methods do not provide the training parameters they used in their papers.

\begin{table}[t]
  \begin{center}
    \resizebox{\columnwidth}{!}{%
		\begin{tabular}{cccc}
			\toprule
      \textbf{Method}                                & \textbf{Dataset}  & \textbf{Damage} & \textbf{Segmentation} \\ \midrule
      \multirow{2}{*}{FC-EF \cite{Daudt_2018}}       & \textit{Split I}  & 0.51 & 0.72 \\
                                                     & \textit{Split II} & 0.43 & 0.73 \\ \hline
      \multirow{2}{*}{FC-Siam-diff \cite{Daudt_2018}}& \textit{Split I}  & 0.52 & 0.73 \\
                                                     & \textit{Split II} & 0.44 & 0.72 \\ \hline
      \multirow{2}{*}{FC-Siam-conc \cite{Daudt_2018}}& \textit{Split I}  & 0.52 & 0.73 \\
                                                     & \textit{Split II} & 0.47 & 0.75 \\ \hline   
      \multirow{2}{*}{Siam-U-Net-Attn-diff}          & \textit{Split I}  & 0.70 & 0.73 \\
                                                     & \textit{Split II} & 0.55 & 0.74 \\ \hline
      \multirow{2}{*}{Siam-U-Net-Attn-conc}          & \textit{Split I}  & 0.69 & 0.73 \\
                                                     & \textit{Split II} & 0.63 & 0.73 \\
			\bottomrule
    \end{tabular}
    }
		\caption{\textbf{Performance Results.} The damage scale classification performance (harmonic means of F1 scores for all damage scales) and the building segmentation F1 scores for the proposed and compared methods using two dataset splitting approaches.}
    \label{table:classification}
  \end{center}	
\end{table}

Table \ref{table:classification} shows a quantified comparison of damage scale classification results for \textit{Split I} and \textit{Split II}.
To evaluate performance, we use the same evaluation metrics as proposed in the xView2 challenge \cite{Gupta_2019}.
The evaluation metric $\text{F1}_s$ for the building segmentation task is defined as:
\begin{align*}
  \text{F1}_s & = \frac{2TP_s}{2TP_s+FP_s+FN_ls}
\end{align*}
where the $TP_s$, $FP_s$, and $FN_s$ are the number of true-positive, false-positive, and false-negative pixels of segmentation results for the entire testing set.
Since the compared methods only produce multi-class damage scale classification masks, we binarize them to create segmentation masks for comparison purposes.

The evaluation metric $\text{F1}_d$ for the damage scale classification task is defined as the harmonic mean of the F1 scores for the four damage scales:
\begin{align*}
  \text{F1}_d & = \frac{4}{\sum_{c\in{\{1,2,3,4\}}}(\text{F1}_c+\epsilon)^{-1}},
\end{align*}
where the $\text{F1}_c$ is the F1 score for the class $c$, which is defined as: 
\begin{align*}
  \text{F1}_c & = \frac{2TP_c}{2TP_c+FP_c+FN_c}.
\end{align*}
The $TP_c$, $FP_c$, and $FN_c$ are the number of true-positive, false-positive, and false-negative pixels of the class $c$ for the testing set.
Note that this testing set does not include background pixels; it only includes pixels from the foreground as determined by the building localization ground truth. 

The proposed approaches outperform the compared methods for the damage scale classification task by a large margin.
With the help of the self-attention module, the proposed methods produce better damage scale classification results using long-range information, as described in Section \ref{sec:method}.
However, \textit{Split II} proves to be more difficult than \textit{Split I}, and we see a drop in performance.
Although there is no overlap between training and testing samples, the degradation indicates that the model might memorize damage levels based on image scenes.
Thus, it could potentially classify patches by recognizing which scene they depict and matching them to scenes already learned, rather than learning to recognize damage in a way that could be applied to new, never-before-seen imagery. 
Therefore, although the two dataset splits are legitimate in terms of separating training and testing data, \textit{Split II} may avoid model overfitting and present a more reliable analysis of model performance. 
For our analysis for the rest of this section, we only consider the results from \textit{Split II}.

All methods achieve similar performance in the building segmentation task.
This is because the proposed methods also use a U-Net for the segmentation task, and the self-attention module that we implement enhances results on the damage scale classification task only. 

The Siamese models in concatenation mode (\ie FC-Siam-conc and Siam-U-Net-Attn-conc models) achieve slightly better results than the model in difference mode (\ie FC-Siam-diff and Siam-U-Net-Attn-diff models).
This is because channel-wise concatenation retains more information than simple differencing.

\begin{figure}[t]
   \begin{center}
     \includegraphics[width=0.7\linewidth]{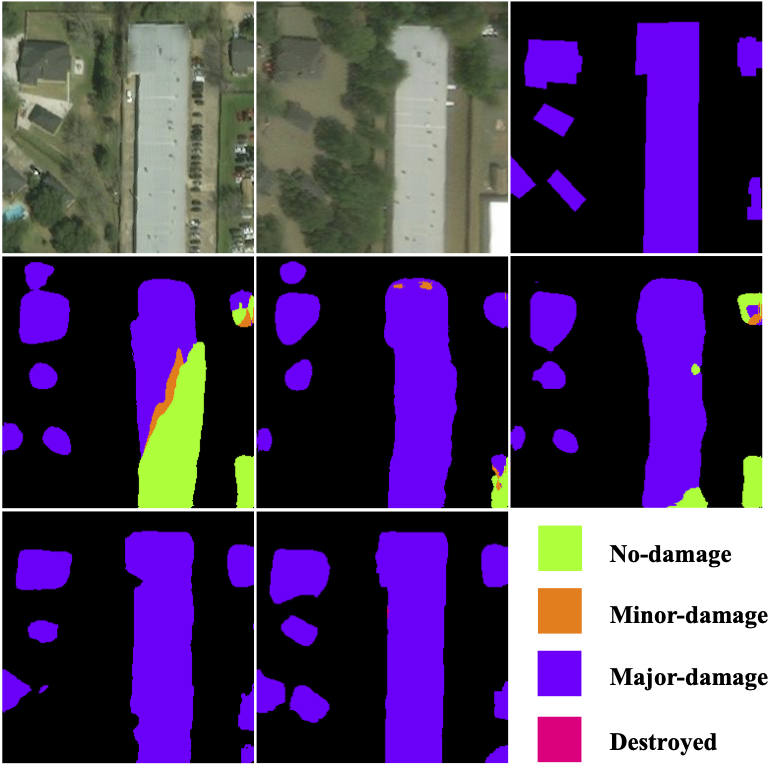}  
   \end{center}
     \caption{\textbf{Comparison of Damage Scale Classification Results using \textit{Split II}}. Top row: pre-event image patch, post-event image patch, and ground truth mask; Second row: FC-EF, FC-Siam-diff, and FC-Siam-conc results; Third row: Siam-U-Net-Attn-diff, and Siam-U-Net-Attn-conc results.}
   \label{fig:compare}
 \end{figure}

Figure \ref{fig:compare} shows the damage scale classification results from the proposed and compared models for a specific scene.
The ground truth classification of the buildings in this example is \textit{major-damage} since a flooding region, which appears as a brownish-yellowish color in the post-event image patch, completely surrounds the buildings.
The results in the second row of Figure \ref{fig:compare} indicate that the three compared methods detect and localize most of the buildings but fail to accurately classify their damage levels.
The FC-Siam-conc model achieves the best results amongst the compared methods.
Compared to the FC-EF and FC-Siam-diff models, it avoids a false alarm detection in the top-left region of the image patch.
Moreover, compared to the FC-Siam-diff model, the concatenation operation from the FC-Siam-conc and FC-EF models helps preserve the necessary information to correctly detect and classify the building in the bottom-left of the image patch.
However, none of these methods assign the correct damage level labels completely.
By comparison, our two proposed methods successfully classify all the buildings, shown in the third row of results of Figure \ref{fig:compare}.
They also segment most of the buildings in the image patch correctly.
Compared to the Siam-U-Net-Attn-diff model, the concatenation operation from the Siam-U-Net-Attn-conc model also helps to correctly detect the building in the bottom-left of the image patch.

\begin{figure*}[t]
  \begin{center}
    \includegraphics[width=0.7\linewidth]{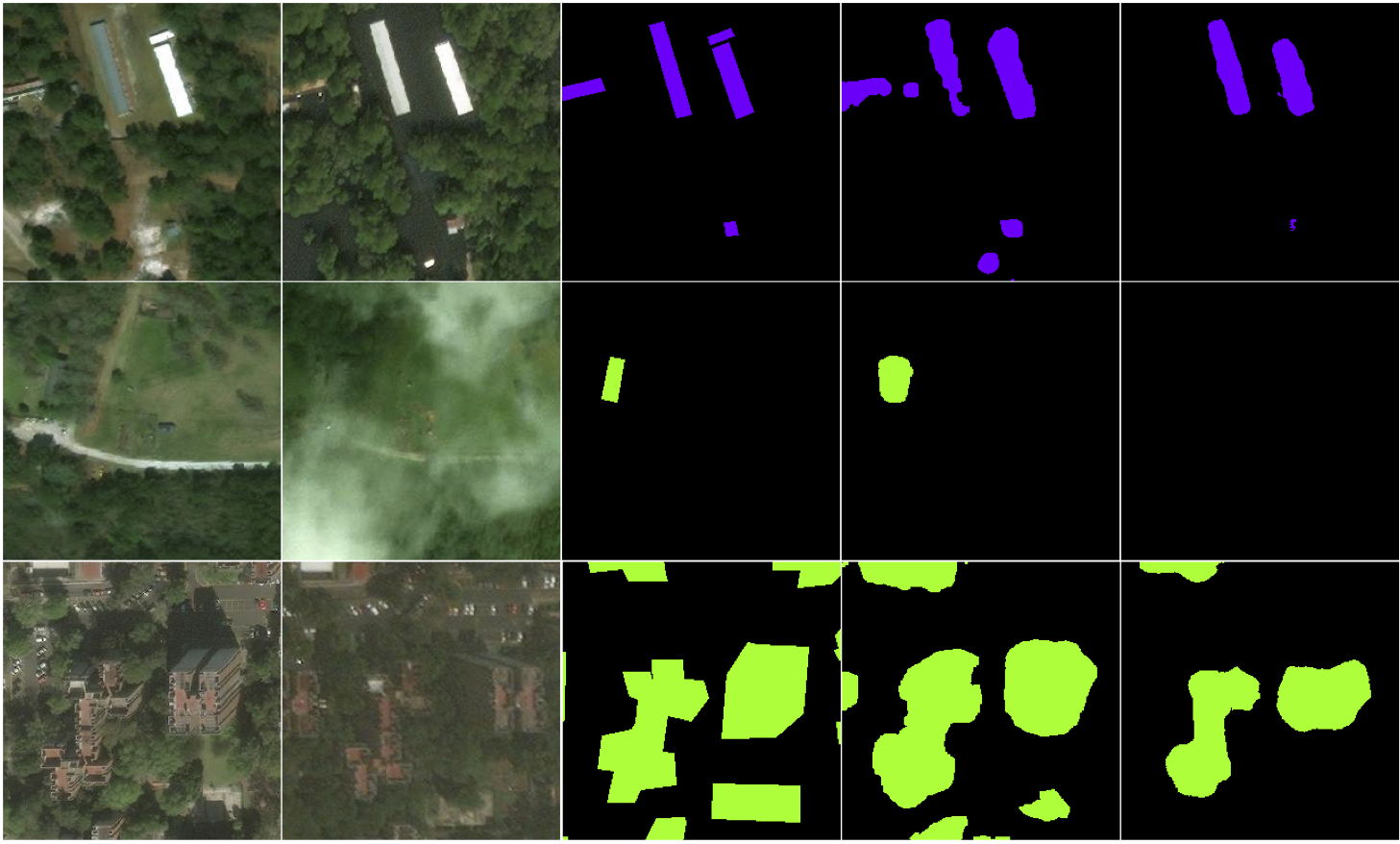}  
  \end{center}
    \caption{\textbf{Robustness of Damage Scale Classification Results}. From left to right: pre-event image patch, post-event image patch, ground truth mask, output of our Siam-U-Net-Attn-conc, and output of FC-Siam-conc \cite{Daudt_2018} using \textit{Split II}. Each row depicts a different scene from the dataset. The proposed method achieves damage scale classification even for image patches with occlusion and different off-nadir angles. }
  \label{fig:compare_more}
\end{figure*}

\begin{figure}[t]
  \begin{center}
    \includegraphics[width=0.8\linewidth]{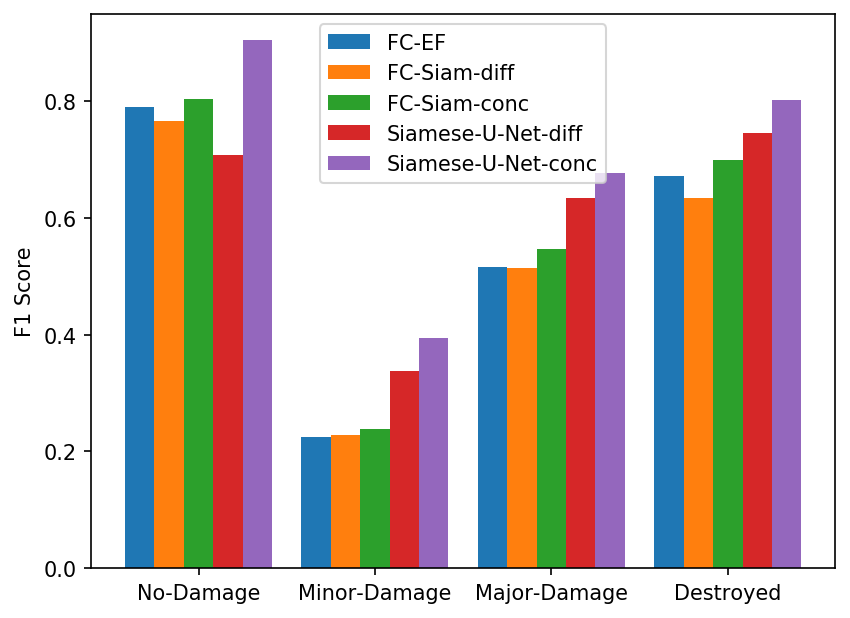}  
  \end{center}
    \caption{\textbf{F1 Scores Based on Damage Scale Level.} These results indicate F1 scores for different damage scales obtained with the compared and proposed methods using \textit{Split II}.}
  \label{fig:per-disaster}
\end{figure}

\begin{figure}[t]
  \begin{center}
    \includegraphics[width=0.7\linewidth]{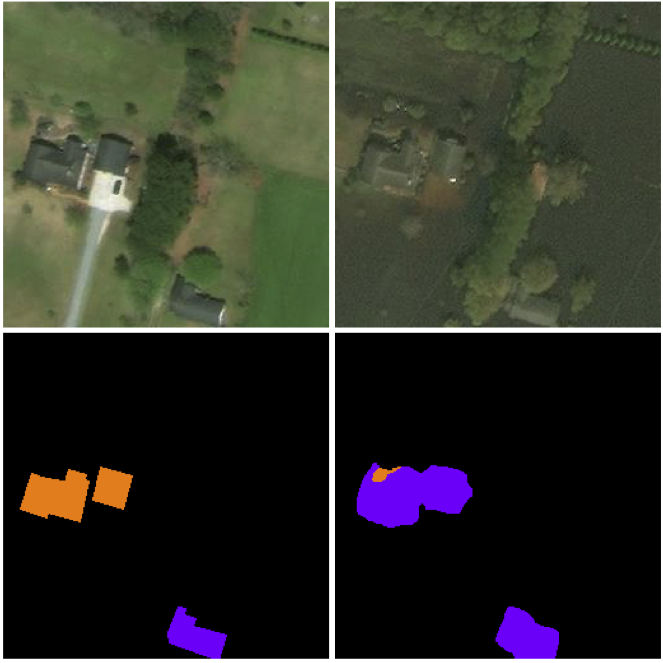}  
  \end{center}
    \caption{\textbf{Failure Case for \textit{Minor-Damage} Building.} From left to right and top to bottom: pre-disaster image patch, post-disaster image patch, ground truth mask, and damage scale classification output mask from Siam-U-Net-Attn-conc model. Buildings labeled as \textit{minor-damage} with the color orange in the ground truth mask are incorrectly classified by the Siam-U-Net-Attn-conc model as \textit{major-damage} with the color purple.}
  \label{fig:failed_case}
\end{figure}

Figure \ref{fig:compare_more} depicts some challenging cases in the testing set and highlights the capabilities of our proposed methods to correctly classify them.
The first challenge case involves two buildings partially occluded by trees in the post-event image  patch (\ie, the left-most and bottom-most buildings).
Although the FA-Siam-conc model accurately classifies the buildings in the center of the image patch, it misses the left building entirely and most of the bottom building.
By comparison, our Siam-U-Net-Attn-conc model detects both of the occluded buildings correctly.
However, it also reports a false alarm.
The second challenge case we consider contains more occlusion, due to cloud cover.
The compared method completely fails to detect the building covered by the clouds.
Even though it is difficult for a human to recognize the building location in this case, our proposed method detects it.
The third challenge case considers two image patches of a co-registered scene taken with different off-nadir angles (\ie, viewing angle from the sensor to the ground).
Both of the compared method and proposed method achieve an accurate classification of most of the buildings.
However, the compared method misses two buildings located towards the bottom of the image patch.
Although the two buildings are still visible in the post-event image patches, the different off-nadir angles change the appearance of the scene quite a bit.
Nonetheless, the proposed model detects the two buildings, even with this large variation in appearance.
Therefore, our proposed methods provide a more robust damage scale classification than the compared methods.

Figure \ref{fig:per-disaster} presents the F1 scores of each damage scale level. 
Overall, the proposed methods, especially the Siam-U-Net-Attn-conc model, perform better than the compared methods for most damage scale levels.
Most of the methods achieve the best performance on buildings with \textit{no-damage} and achieve the worst performance on buildings with \textit{minor-damage}.
\textit{Minor-damage} buildings present the most difficult challenge because they usually do not exhibit visible damage on the buildings themselves.
Damage assessment experts from~\cite{Gupta_2019} consider buildings as \textit{minor-damage} due to flooding regions, volcano flow, or burned trees partially surrounding them.
This is very similar to the \textit{major-damage} classification except that a building classified as \textit{major-damage} indicate that such elements completely surround that particular building. 
Thus, these two similar damage scale levels present a greater challenge for damage scale classification models.
As shown in Figure \ref{fig:failed_case}, the proposed Siam-U-Net-Attn-conc model fails to recognize that the water region (\ie, the dark green region in the post-event image patch) only partially surrounds the buildings.
Instead, it mislabels it as \textit{major-damage}.
All of the compared and proposed methods demonstrate this behavior, performing worse on the \textit{minor-damage} buildings as compared to other damage scale levels.

\begin{figure}[t]
  \begin{center}
    \includegraphics[width=0.8\linewidth]{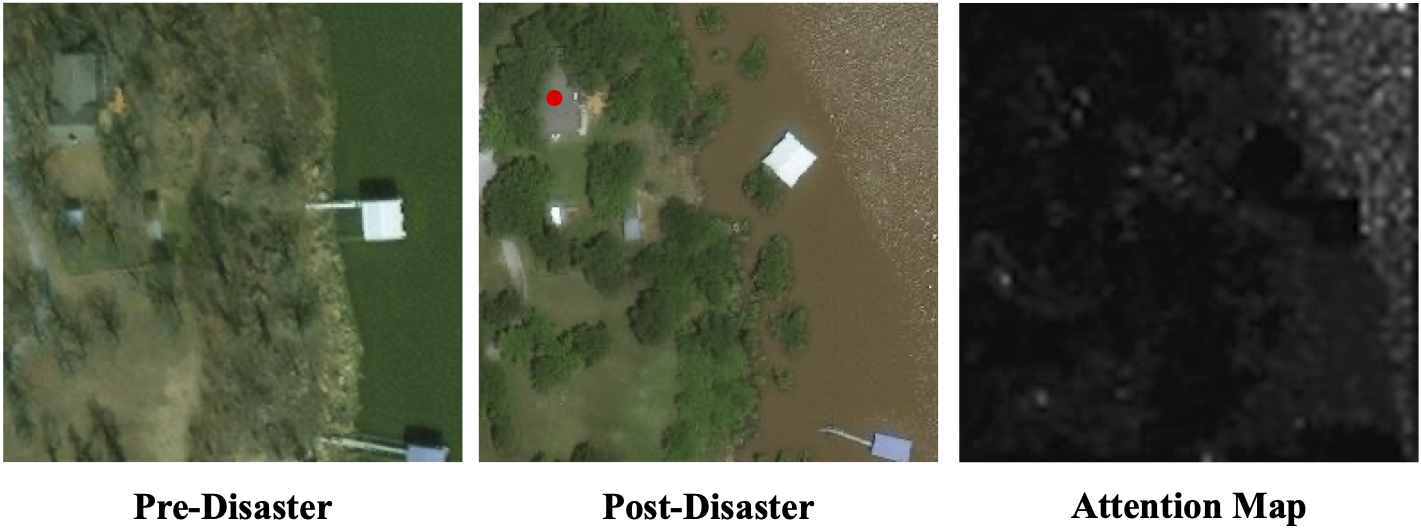}  
  \end{center}
    \caption{\textbf{Attention Map}. From left to right: pre-event image patch, post-event image  patch with query point (\ie, the red point), and attention map associated with the given query point. Brighter regions in the attention map signify greater importance of those pixels to the classification of the query point.}
  \label{fig:attention}
\end{figure}

The utility of the self-attention module can also be visualized.
We portray an attention map in Figure \ref{fig:attention} to demonstrate the effectiveness of the self-attention module.
For a given query location (\ie, the red point in the post-event image patch), we can obtain the corresponding attention map.
Pixel values in the attention map indicate the importance of that pixel to the query point.
The brighter a pixel is, the more important it is to the classification efforts at the query point.
In the area shown in the example, the brownish-yellowish area in the post-event image patch indicates the flooding region.
According to the attention map, the self-attention model highlights this flooding area, which aids the model in classifying the buildings' damage levels.

\begin{figure}[t]
  \begin{center}
    \includegraphics[width=0.9\linewidth]{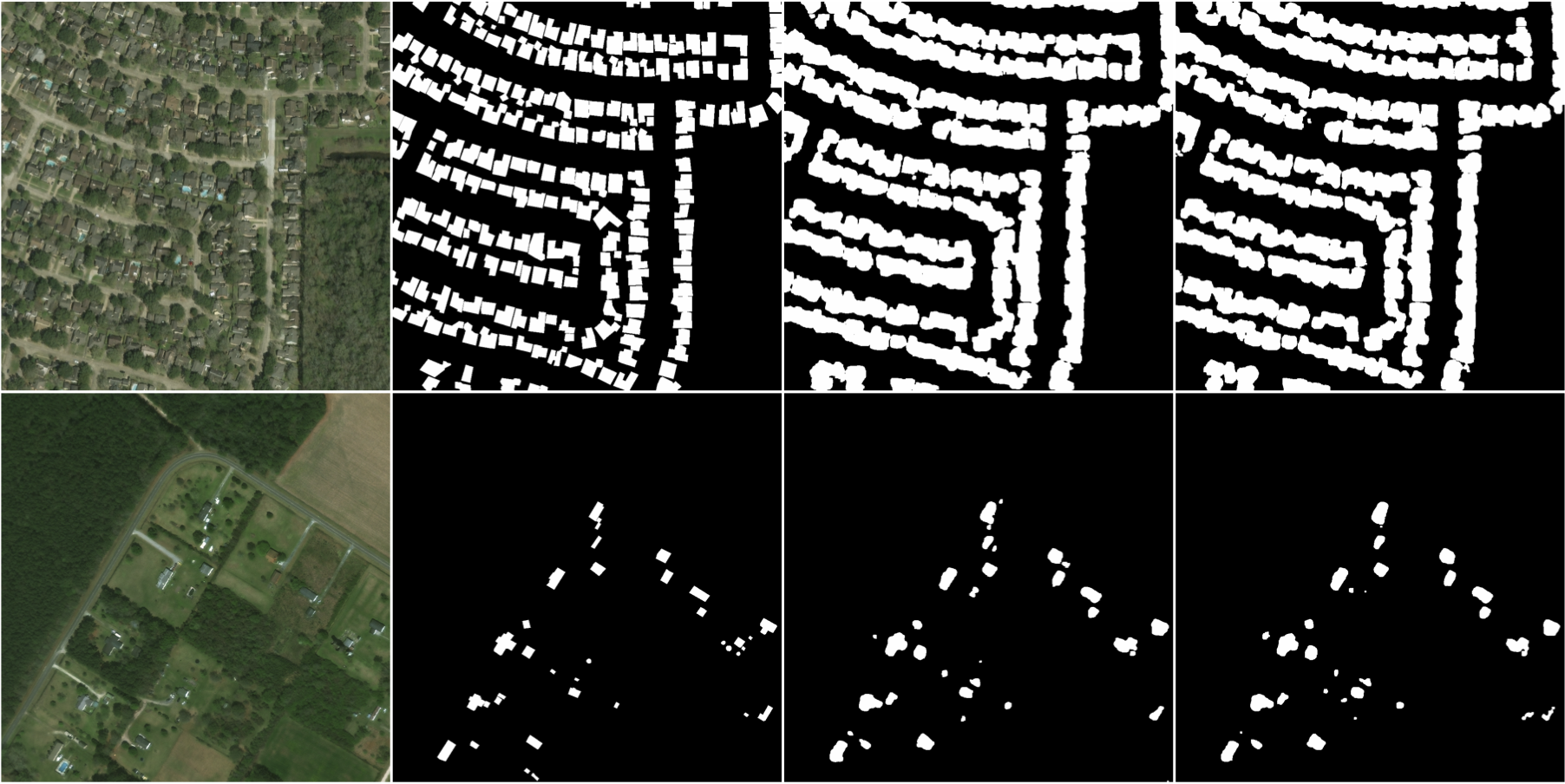}  
  \end{center}
    \caption{\textbf{Full-Resolution Building Segmentation Masks}. From left to right: original image, ground truth mask, Siam-U-Net-Attn-diff result, and Siam-U-Net-Attn-conc result. Each row depicts a different scene from the dataset.}
  \label{fig:segmentation}
\end{figure}

Figure \ref{fig:segmentation} shows some examples of final, full-resolution building segmentation masks constructed from the image patches used by the proposed models.
Since the models operate on image patches, the model results must be stitched together to create a full-resolution mask corresponding to the original image.
For visualization purposes, we crop the original full-resolutions images differently than the method outlined previously. 
This cropping method is only performed on images in the testing dataset, solely for the purpose of producing better and more coherent visual results.
The goal of this different procedure is to reduce abrupt edges at the boundaries of adjacent patches.
We use a moving-window approach to crop full-resolution images into patches with overlapping regions.
The stride for the moving-window is 32 pixels in both the vertical and horizontal directions. 
Then, the model analyzes these patches and produces corresponding segmentation maps.
Next, we use a voting strategy for each pixel contained in the overlapping regions to determine the final segmentation mask.
More specifically, we sum the probabilities of each class to calculate five overall probabilities that a specific pixel belongs to each of the damage level classes.
Then, we label the pixel under consideration as the class with the maximum probability.
The two examples in Figure \ref{fig:segmentation} show that our proposed methods perform well on the building segmentation task in cases with dense and sparse building densities.

\section{Conclusion}
\label{sec:conclusion}

In this paper, we present a Siam-U-Net-Attn model with self-attention for building segmentation and damage scale classification in satellite imagery.
The proposed technique compares pairs of images captured before and after disasters to produce segmentation masks that indicate damage scale classifications and building locations.
Results show that the proposed model accomplishes both damage classification and building segmentation more accurately than other approaches with the xView2 dataset.
We use the self-attention module to enhance damage scale classification by considering information from the entire image.

{\small
\bibliographystyle{ieee_fullname}
\bibliography{refs}
}

\end{document}